\PassOptionsToPackage{table}{xcolor}
\documentclass[conference]{IEEEtran}
\IEEEoverridecommandlockouts
\usepackage{cite}
\usepackage{listings}
\usepackage{amsmath,amssymb,amsfonts}
\usepackage{algorithmic}
\usepackage{graphicx}
\usepackage{svg}
\usepackage{textcomp}
\usepackage[table]{xcolor}
\usepackage{colortbl}
\usepackage{acronym}
\usepackage{amssymb}
\usepackage{subcaption}

\usepackage{comment}
\newboolean{showcomments}
\setboolean{showcomments}{true}
\ifthenelse{\boolean{showcomments}}
{ \newcommand{\mynote}[3]{
     \fbox{\bfseries\sffamily\scriptsize#1}
        {\small$\blacktriangleright$\textsf{\emph{\color{#3}{#2}}}$\blacktriangleleft$}}}
{ \newcommand{\mynote}[3]{}}
\definecolor{asparagus}{rgb}{0.53, 0.66, 0.42}

\def\BibTeX{{\rm B\kern-.05em{\sc i\kern-.025em b}\kern-.08em
    T\kern-.1667em\lower.7ex\hbox{E}\kern-.125emX}}
\begin{document}
\begingroup
\let\section\relax
\let\chapter\relax
\acrodef{ML}{Machine Learning}
\acrodef{SoC}{System-on-Chip}
\acrodef{PU}{Processing Unit}
\acrodef{CPU}{Central Processing Unit}
\acrodef{GPU}{Graphics Processing Unit}
\acrodef{LLM}{Large Language Model}
\acrodef{AOT}{Ahead-of-Time}
\acrodef{IR}{Intermediate Representation}
\acrodef{MLIR}{Multi-Level Intermediate Representation}
\acrodef{DSE}{Design Space Exploration}
\acrodef{PE}{Processing Element}
\acrodef{KV}{Key--Value}
\acrodef{API}{Application Programming Interface}
\acrodef{FP16}{16-bit Floating Point}
\acrodef{INT8}{8-bit Integer}
\acrodef{LLVM}{Low Level Virtual Machine}
\acrodef{SPIR-V}{Standard Portable Intermediate Representation -- Vulkan}
\acrodef{DSE}{Design Space Exploration}
\acrodef{SD}{Speculative Decoding}

\endgroup
\title{Compiler-Assisted Speculative Sampling for Accelerated LLM Inference on Heterogeneous Edge Devices}

% \author{\IEEEauthorblockN{Alejandro Ruiz y Mesa$^{*}$}
% \IEEEauthorblockA{
% \textit{TU Dresden}\\
% alejandro.ruiz\_y\_mesa\\
% @mailbox.tu-dresden.de\\
% Dresden, Germany}
% \and
% \IEEEauthorblockN{ Guilherme Korol}
% \IEEEauthorblockA{\textit{NXP Semiconductors}\\
% guilherme.korol@nxp.com\\
% Hamburg, Germany }
% \and
% \IEEEauthorblockN{Moritz Riesterer}
% \IEEEauthorblockA{\textit{NXP Semiconductors}\\
%  moritz.riesterer@nxp.com\\
% Munich, Germany}

% \and
% \IEEEauthorblockN{João Paulo C. de Lima}
% \IEEEauthorblockA{\textit{TU Dresden, ScaDS.AI}\\
% joao.lima@tu-dresden.de\\
% Dresden, Germany}

% \and
% \IEEEauthorblockN{Jeronimo Castrillon}
% \IEEEauthorblockA{\textit{TU Dresden, ScaDS.AI}\\
% jeronimo.castrillon@tu-dresden.de\\
% Dresden, Germany}

\author{%
    \IEEEauthorblockN{Alejandro Ruiz y Mesa$^{*}$\IEEEauthorrefmark{2}, Guilherme Korol\IEEEauthorrefmark{3}, Moritz Riesterer\IEEEauthorrefmark{3}, \\
    Jo\~ao Paulo C. de Lima\IEEEauthorrefmark{2}\IEEEauthorrefmark{4}, Jeronimo Castrillon\IEEEauthorrefmark{2}\IEEEauthorrefmark{4}}
    \IEEEauthorblockA{\IEEEauthorrefmark{2}Dresden University of Technology, Dresden, Germany\\
        Email: alejandro.ruiz\_y\_mesa@mailbox.tu-dresden.de, \{joao.lima, jeronimo.castrillon\}@tu-dresden.de}
    \IEEEauthorblockA{\IEEEauthorrefmark{3}NXP Semiconductors, Germany\\
        Email: \{guilherme.korol, moritz.riesterer\}@nxp.com}
    \IEEEauthorblockA{\IEEEauthorrefmark{4}Center for Scalable Data Analytics and Artificial Intelligence (ScaDS.AI), Dresden, Germany}
    \thanks{$^{*}$Work realized during internship at NXP Semiconductors.}
}

\maketitle

\begin{abstract}
%The growing demand to deploy LLMs on resource-constrained edge devices makes reducing inference latency a central priority. 
LLM deployment on resource-constrained edge devices faces severe latency constraints, particularly in real-time applications where delayed responses can compromise safety or usability. 
Among many approaches to mitigate the inefficiencies of sequential token-by-token generation, 
% JC: will "autoregressive generation" be clear at this point?
% JP: I think for a general ML HW and ML compiler audience is fine. I replaced with a simpler term.
Speculative Decoding (SD) has emerged as a promising technique.
However, SD at the edge is hindered by two major challenges: (1) integrating SD into a compiler-based workflow without sacrificing performance or programmability, and (2) exploiting the heterogeneous compute resources of modern SoCs through carefully designed partitioning strategies.
%Although \ac{SD} is a promising technique for mitigating the sequential bottleneck of autoregressive generation, its effective use at the edge is hindered by two major problems: integrating a generative pipeline with \ac{SD} into a compiler workflow without sacrificing performance or programmability, and exploiting the heterogeneous compute resources of modern SoCs through partitioning strategies.
This work addresses these challenges by using an analytical cost model that explores heterogeneous hardware configurations and guides coarse-grained partitioning of LLM subgraphs, particularly with edge-typical short input sequence lengths. The cost model predicts when speculative sampling and heterogeneous execution are jointly beneficial and is validated on an edge device featuring a hexacore Cortex-A CPU and a Mali GPU, revealing up to 1.68$\times$ speedup for translation tasks, closely matching analytic expectations.
\end{abstract}


% =====================================================================
\section{Introduction}
% =====================================================================

Enabling \acp{LLM} on edge devices is essential for applications requiring privacy, robustness, and low latency, such as offline health assistants or real-time translation~\cite{zheng_review_2025}. However, deploying \acp{LLM} on edge platforms faces severe challenges under the scarce compute, power, and memory available. Modern \ac{SoC} platforms integrate multiple heterogeneous \acp{PU}, such as multi-core CPUs, mobile GPUs, and accelerators, creating opportunities for workload acceleration. Still, transformer-based \acp{LLM} present a particularly challenging workload due to their autoregressive decoding, which generates one token at each forward pass and imposes hard sequential dependencies. Addressing these challenges, therefore, requires a combination of algorithmic-, system-, and hardware-level optimization techniques. 
%\jp{I changed the flow by moving motivation -> constraints -> hardware reality -> remaining workload challenge -> implication}

\ac{SD}, in particular, has demonstrated remarkable improvements in the generative pipeline on high-end GPUs, achieving speedups up to 6.5$\times$ compared to traditional autoregressive decoding~\cite{leviathan_fast_2023, li_eagle-3_2025}. \ac{SD} converts savings from traditional algorithmic optimizations (e.g., quantization, pruning, and distillation) into end-to-end latency reductions by relying on an inexpensive draft model to reduce the number of costly target-model decoding steps, with performance determined by the draft cost, verification cost, and acceptance rate. This approach naturally lends itself to migrating LLM inference from cloud to edge platforms. 
% Despite the limited processing power of edge devices, this algorithmic optimization can offer a substantial performance boost with minimal hardware overhead, requiring only a slight increase in transistor count.

A key question is how to integrate \ac{SD} into the software stack of an edge system in a way that preserves productivity for developers while enabling meaningful hardware-aware optimization. Without such careful integration, \ac{SD} is not inherently beneficial: naive adoption or poor hardware mapping can negate its advantages and even increase end-to-end latency.
% \gk{I think we are missing a motivation. Did we have one before? Anyway, one that we could use is to say that even though spec dec is being (wudely?) used [some ref], it can even slow down generation if used naively or wrongly mapped (see results in sec ABC).}\jp{See my suggestion}
In this work, we build upon IREE\cite{linux_foundation_iree_2025}, an open-source ML compiler to provide a unified representation of \ac{SD} that supports explicit device affinities, coarse-grained heterogeneous mapping, and \ac{AOT} compilation across multiple backends. On top of that, we leverage the analytical cost model from~\cite{leviathan_fast_2023} to guide \ac{LLM} partitioning and mapping across heterogeneous \acp{PU}. 
% JC: Suggest to use an acronym package for consistency. 
Concretely, our contributions are the following:

\begin{itemize}
    \item We apply an analytical cost model to (i) determine when to use speculative sampling and to (ii) map it 
    % perform spatial mapping 
    onto heterogeneous \acp{PU} for efficient \ac{LLM} inference acceleration;
    % (Section~\ref{subsec:dse}).
    \item We present compiler-level abstractions of speculative sampling that enable device placement decisions for heterogeneous edge execution, comparing monolithic and modular compilation strategies; 
    % (Section~\ref{subsec:abstractions}).
    \item We demonstrate that our optimized heterogeneous mappings can outperform homogeneous CPU execution by up to $1.68\times$ by reducing speculation overhead;
    % , though the effectiveness is highly hardware-dependent and must be evaluated for each system configuration (Section~\ref{subsec:cost_coeff}).
    \item We validate the proposed cost model using a hexacore Cortex-A55 CPU and a Mali-G310 GPU integrated on an NXP i.MX95 SoC, for translation tasks, with a $4\%$ deviation from analytical expectations. % (Section~\ref{sec:val_and_dis}).
    % JC: Say somethign about how good/acurate/fast/... it is
    % \item We provide insights on the trade-offs between monolithic and modular compilation strategies for expressing generative pipelines.
    % JC: Unsure about the real value of this contribution. Is this like a guideline that other people can follow? 
\end{itemize}
% JC: Section III has a way too generic title "Methodology". I'd rather have titles that are more descriptive. It would be great if there is a simple mapping from the contribs above to sections/subsections in the paper. If you manage to do that, you should insert fwd-refs in the bullet list above. 

% Together, these contributions demonstrate how algorithmic techniques and compiler-level abstractions can jointly improve the feasibility of \ac{LLM} deployment on heterogeneous edge hardware.
% JC: Can this be supported by some quantitative results from the paper? 

% =====================================================================
\section{Background and Related Work}
\label{sec:background_related}
This section builds on insights into computational bottlenecks inherent in \ac{LLM} inference, \ac{SD} as an algorithmic optimization technique, and the role of ML compilers in enabling efficient deployment across diverse hardware platforms. Toward the end, we review related work on accelerating neural network inference and highlight how our approach differs from existing techniques.

% \subsection{Heterogeneous Edge Computing}

% Modern edge SoCs combine general-purpose CPUs with GPUs and increasingly dedicated NPUs or other DSAs. CPUs on edge devices offer the programmability and control performance necessary for general-purpose workloads, but their limited vector width and low clock frequencies constrain throughput for transformer-based models. Mobile GPUs, such as ARM Mali or Qualcomm Adreno architectures, provide greater parallel compute density but operate under tight energy and thermal envelopes. These GPUs employ relatively narrow Single Instruction, Multiple Threads (SIMT) execution units originally designed for graphics workloads, where compute dispatch typically occurs through graphics APIs such as OpenCL or Vulkan~\cite{lee_-device_2019, jiang_profiling_2020}. 

\subsection{LLM Inference Bottlenecks}
\label{subsec:llm_bottlenecks}
Inference in transformer-based \acp{LLM} proceeds in two phases, each presenting distinct computational challenges~\cite{yuan_llm_2024}. The prefill phase processes the entire input sequence in parallel, populating the \ac{KV} cache across all transformer layers. In contrast, the decode phase generates tokens sequentially, with each token depending on the previous model state. As such, opportunities for parallelism are limited. Furthermore, the hidden dimension of the \ac{LLM}, denoted as $d$, plays a pivotal role in shaping the computational characteristics of the model~\cite{kim_full_2023}. The sequence length ($S_L$) is often categorized relative to $d$: short sequences ($S_L \ll d$) are dominated by the linear layers, whereas the attention layers dominate in long sequences ($S_L \gg d$). This distinction is crucial for understanding performance bottlenecks.

\subsection{Speculative Decoding}
\label{subsec:spec_dec}

\begin{figure}[b]
    \centering
    \includegraphics[width=0.9\linewidth]{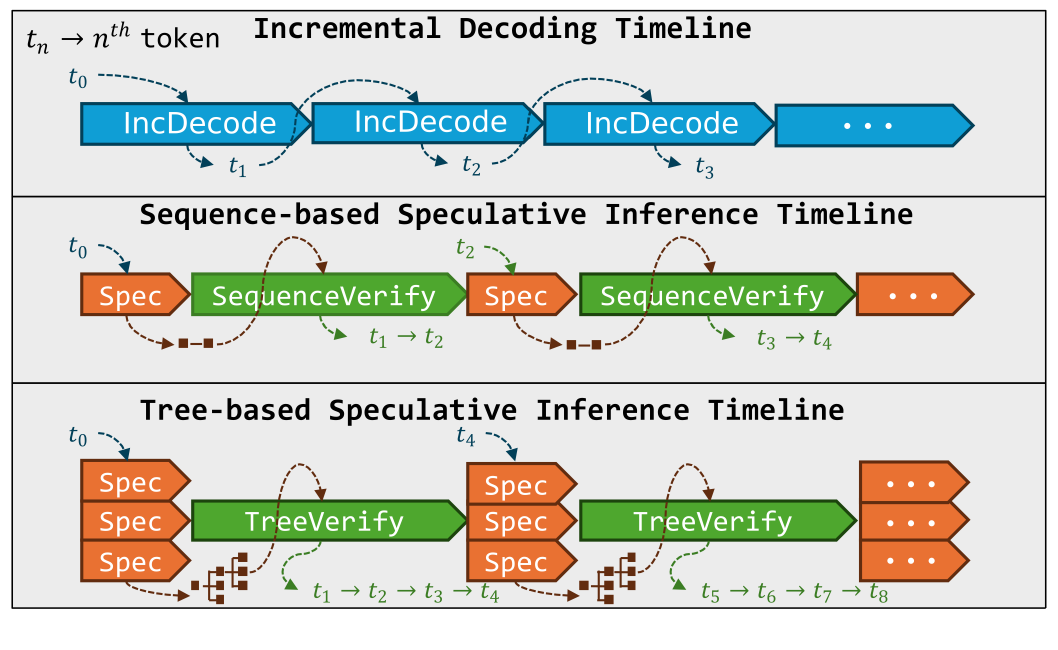}
    \caption{Three generative pipelines: standard sampling (top, in blue), and two variants of \ac{SD}: sequential drafting (center) and tree-based drafting (bottom). Adapted from~\cite{miao_specinfer_2024}. 
    % \gk{we just noticed that this fig is used as-is from [15]. We must re-draw it to use (ideally following the style used in the other figs)!}
    }
    \label{fig:specdec_overview}
\end{figure}

\ac{SD} accelerates autoregressive generation by reducing the number of full forward passes needed from the \ac{LLM} (referred to as the target model). The process is conceptually divided into two phases. In the \textit{speculation phase}, the system generates a sequence of candidate tokens using a computationally inexpensive mechanism. This may be a smaller transformer, a distilled or pruned version of the original network, or a specialized predictive module, and is designed to approximate the next-token distribution of the target model at a significantly lower computational cost. In the subsequent \textit{verification phase}, the larger target model evaluates the drafted tokens in a single forward pass, in a parallel way similar to the prefill phase of the \ac{LLM} inference. Speculated tokens are accepted or rejected according to a selection criterion. 
Fig.~\ref{fig:specdec_overview} illustrates standard incremental decoding (top) and two speculative strategies: sequence-based (middle) and tree-based (bottom). Orange segments indicate speculation by a lightweight model, while green segments denote verification by the target model. 

% Two different speculative pipelines are shown in Fig.~\ref{fig:specdec_overview}. At the top, the Incremental Decoding Timeline depicts the standard sequential execution of the decode phase without speculative techniques. The middle and bottom timelines present two \ac{SD} strategies: Sequence-based (speculative sampling) and Tree-based Speculative Inference Pipelines. In both cases, the orange segments stand for the speculative generation performed by a lightweight model, the small orange squares represent a speculated token, while the green segments denote the validation phase carried out by the target model, which accepts or rejects the drafted tokens.

Speculative sampling~\cite{leviathan_fast_2023} is a form of \ac{SD} that avoids the training of a drafter mechanism, since it employs a smaller transformer model as the speculation strategy. The structural similarity between drafter and target yields sufficiently correlated logits to produce meaningful acceptance rates, enabling substantial speedups on server hardware. On edge devices, however, acceptance behavior is more complex. Quantization alters the relative probabilities of the vocabulary distribution, resulting in acceptance rates lower than those observed with full-precision models~\cite{zhang_speculative_2025}.

In speculative sampling, the achieved speedup depends not only on the alignment between the drafter and target model distributions, but also on the workload, the task, and the hardware-software environment in which the accelerated \ac{LLM} is executed. A formalism modeling the hardware impact on the speculative sampling speedup is developed in~\cite{leviathan_fast_2023}. It presents the speedup $S$ as a function of the acceptance rate's expected value ($\alpha$, the mean proportion of accepted tokens), the speculated draft length ($\gamma$, the sequence length of the speculated tokens), and the cost coefficient $c = t_{\text{draft}}/t_{\text{target}}$, a hardware- and software-dependent parameter representing the ratio between the latency of a single drafter forward pass ($t_{\text{draft}}$) and that of the target model ($t_{\text{target}}$). The speedup $S$ is given by~\eqref{eq}.

\begin{equation}
S\left(\alpha, \gamma, c\right) = \frac{1-\alpha^{\gamma+1}}{(1 - \alpha)(\gamma c + 1)}\label{eq}
\end{equation}

Moreover, the condition $c < \alpha$ must hold to achieve any speedup at all. This guarantees that there is at least one value of $\gamma$ that yields a speedup greater than 1. Furthermore, the optimal draft length $\gamma^*$ that maximizes the speedup depends on both the hardware configuration and the quality of the drafting mechanism~\cite{leviathan_fast_2023}.

\begin{figure*}[th]
    \centering
    \begin{subfigure}[b]{0.72\textwidth}
        \centering
        \includegraphics[width=\textwidth]{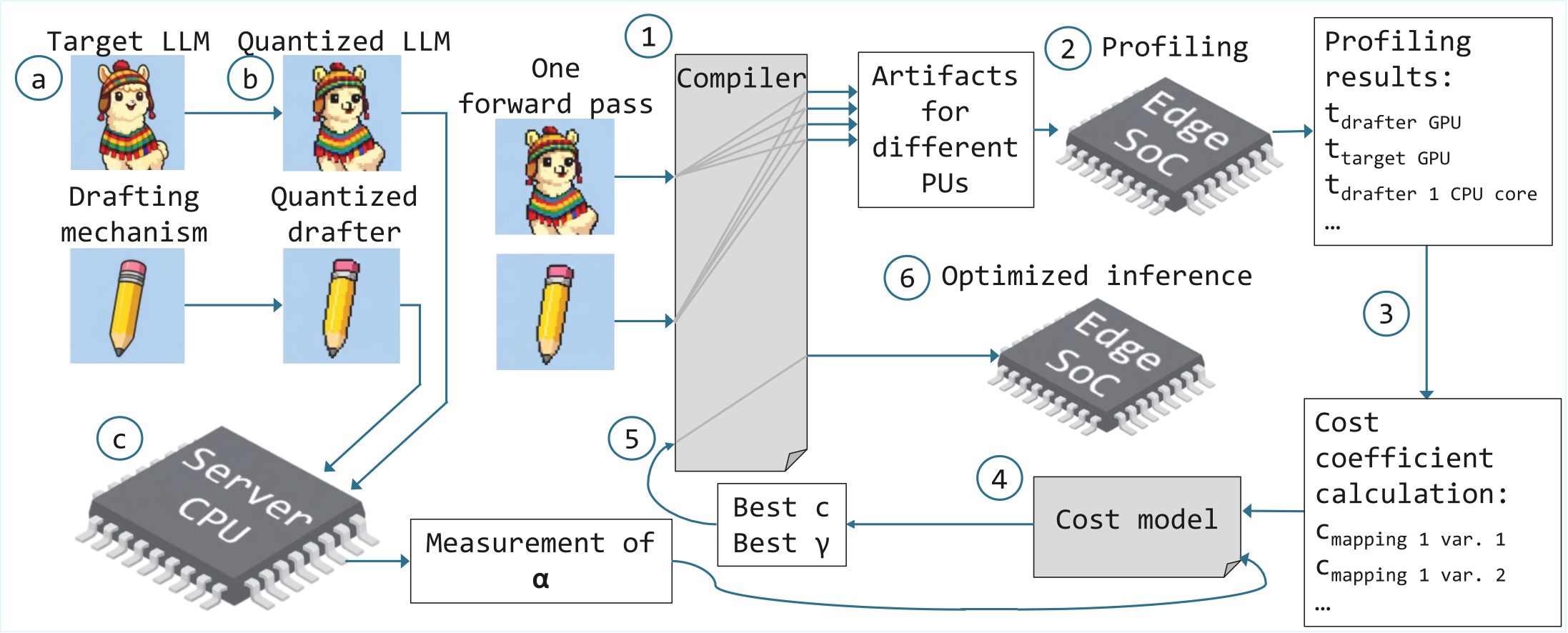}
        \caption{Overview of the optimization flow}
        \label{fig:methodology_overview}
    \end{subfigure}
    \hfill
    \begin{subfigure}[b]{0.25\textwidth}
        \centering
        \includegraphics[width=\textwidth]{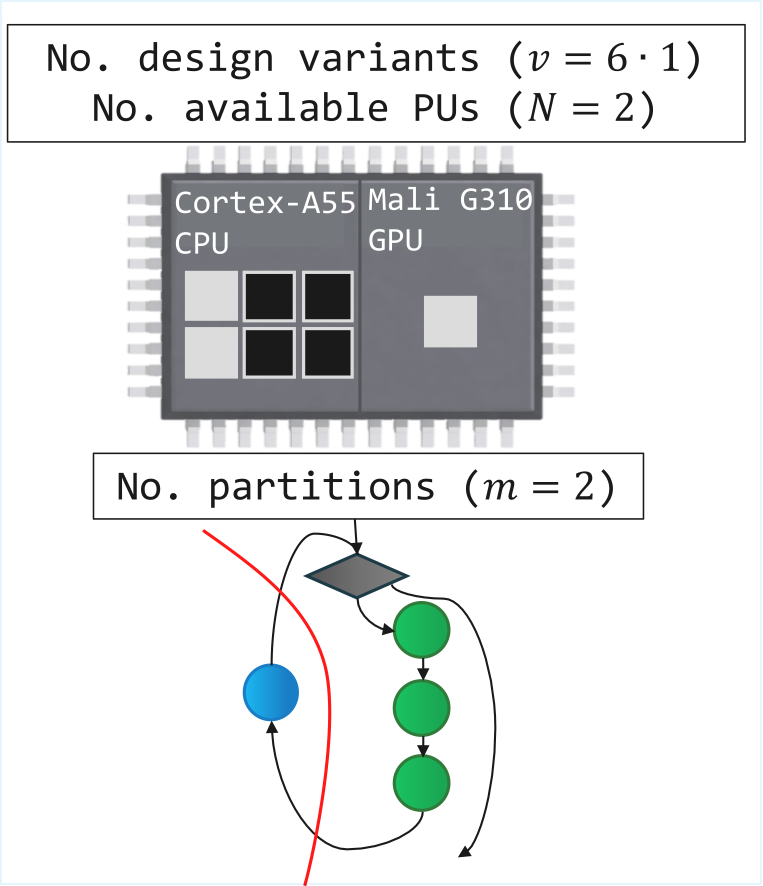}
        \caption{Example of design space
        % with $v = 6$ design variants and two available \acp{PU} (top) and $m = 2$ subgraph partitions (bottom). The CPU provides six cores and the GPU a single shader, but in this example only two CPU cores (in grey) are assumed to be available for mapping.
        }
        \label{fig:dse}
    \end{subfigure}
    \caption{Overview of heterogeneous mapping workflow for speculative sampling on edge devices.
    %\jp{Stick to one single font style. I'd suggest adopting Consolas for all hand-drawn figures, but it's not a must.}
    % \jp{There's plenty of space in this figure. I suggest merging Fig. 3 here.}
    }
    \label{fig:methodology_all}
\end{figure*}

\subsection{Related Work}
\label{subsec:related}
Traditional algorithmic optimizations (e.g., quantization, pruning, and knowledge distillation) reduce compute and memory requirements, effectively producing a cheaper draft model and/or improving draft–target agreement.
Recent systems (e.g., Sequoia~\cite{chen_sequoia_2024}, DuoDecoding~\cite{lv_duodecoding_2025}, and Dovetail~\cite{zhang_dovetail_2024}) further show that the optimal draft/verifier configuration depends on hardware characteristics and device heterogeneity, motivating joint algorithm–system co-design.
%Research on accelerating neural network inference spans algorithmic- and system-level techniques. Traditional algorithmic optimizations (e.g., quantization, pruning, and knowledge distillation) reduce compute and memory requirements on constrained devices~\cite{deng_model_2020}. 
%\ac{SD} methods and their interaction with system characteristics have also been explored: 
%Sequoia adapts draft tree structures to hardware features~\cite{chen_sequoia_2024}, DuoDecoding schedules speculative tasks across heterogeneous devices~\cite{lv_duodecoding_2025}, and Dovetail partitions speculative workloads between CPU and GPU~\cite{zhang_dovetail_2024}. 
However, these approaches primarily target high-end GPU platforms rather than edge-grade SoCs. Complementary work on mapping neural workloads onto heterogeneous compute resources, including polyhedral optimization and design-space exploration, appears in frameworks like Tiramisu~\cite{baghdadi_tiramisu_2018}, ZigZag~\cite{mei_zigzag_2020}, and FlexInfer~\cite{na_flexinfer_2025}. Related efforts on the edge explore algorithm-hardware co-design on reconfigurable platforms, combining algorithmic optimizations, such as pruning and early-exit~\cite{korol_date_2023, korol_isvlsi_2023}. 
% JC: Should cite Guilhermes work in DATE and ISVLSI? 

Neural network multi-tenancy represents another relevant research direction. Frameworks like MAGMA~\cite{kao_magma_2022} and Adyna~\cite{li_adyna_2025} address how concurrent models contend for limited compute and memory resources. Such systems reveal structural similarities to speculative sampling pipelines, where drafter and target models form a multi-tenant workload sharing limited edge resources. However, these works focus on the orchestration of independent models, rather than coupled models within a pipeline. Furthermore, they evaluate against accelerator models or simulators rather than on silicon, as done in this work.

Our work differs since (1) we target an edge \ac{SoC} rather than high-end GPU platforms, addressing the unique constraints of resource-limited devices; (2) we provide the first exploration of coarse-grained heterogeneous partitioning strategies for speculative sampling, leveraging high-level compiler frontend abstractions to express device placement decisions; (3) we apply an analytical cost model that jointly optimizes algorithmic acceleration (speculative sampling) and hardware mapping decisions for heterogeneous edge execution; and (4) we validate our approach on real silicon rather than relying on simulators, providing empirical evidence of the method accuracy.

% =====================================================================
\section{Heterogeneous Mapping Framework for Speculative Sampling}
\label{sec:methodology}

% JC: A similar intro to Section II would be good as well. 

This section presents a systematic approach to evaluate mappings of \acp{LLM} with speculative sampling on edge devices. It begins by providing an overview of the complete compilation and optimization workflow in Sec.~\ref{subsec:workflow_overview}, followed by the formulation of the heterogeneous mapping problem as a \ac{DSE} in the Sec.~\ref{subsec:dse} and~\ref{subsec:measurement}. Sec.~\ref{subsec:abstractions} details how abstractions can be aligned with hardware-aware execution to preserve productivity while enabling good performance.

\subsection{Overview of the Compilation and Optimization Workflow}
\label{subsec:workflow_overview}

Our approach combines offline profiling, analytical modeling, and compiler-assisted code generation (Fig.~\ref{fig:methodology_overview}). We evaluate speculative sampling~\cite{leviathan_fast_2023} for its training-free nature, though the workflow is generalizable to other sequential \ac{SD} techniques. Our workflow consists of the following steps:

\paragraph{Offline quantization (Sec.~\ref{subsec:measurement})} Model quantization is performed offline. In Fig.~\ref{fig:methodology_overview}, from \textcircled{a} to \textcircled{b} the target model and the drafter model are quantized with different schemes that match the available arithmetic on the target \ac{SoC}. In \textcircled{c}, the acceptance rate $\alpha$ is measured for the different combinations of quantized target-drafter mechanisms.
\paragraph{Profiling (Sec.~\ref{subsec:dse} and~\ref{subsec:measurement})} We compile forward passes of both models targeting all \acp{PU} \textcircled{1}, profile them on hardware to measure $t_{\text{draft}}$ and $t_{\text{target}}$ \textcircled{2}, and calculate the cost coefficient $c = t_{\text{draft}}/t_{\text{target}}$ for each design variant \textcircled{3} (Fig.~\ref{fig:dse}).
\paragraph{Exploration (Sec.~\ref{subsec:dse})} With $\alpha$ and $c$ values in hand, the analytical cost model \textcircled{4} is evaluated to determine the optimal draft length $\gamma$ and device mapping for each hardware configuration (see \textcircled{5}). 
%This \ac{DSE} loop explores the space of heterogeneous \ac{PU} assignments, identifying configurations where \ac{SD} yields speedup $S > 1$.
\paragraph{Compilation and Inference (Sec.~\ref{subsec:abstractions})} The complete \ac{SD} pipeline is compiled and executed using IREE in \textcircled{6}.

\subsection{Design Space Encoding, Exploration, and Evaluation}
\label{subsec:dse}
% JC: This subsec seems to jump directly into the detail. I usually prefer if this sort of sections start with a graphical top-level view of the methodology, its steps, relations, and so on. And then have subsection talk about the details. That is: First big pic, then details. What goens in, what steps does it undergo, what goes out. In a "block diagram" like that it is easier to highlight the contribs. 

We formulate the heterogeneous execution of an \ac{LLM} with speculative sampling as a \ac{DSE} problem over static spatial mappings of the computational graph: each subgraph is assigned ahead of time to a \ac{PU} and no dynamic remapping is considered. Additionally, the search concentrates on coarse-grained partitions that separate the drafter (speculation mechanism model) from the target model (\ac{LLM} to be accelerated), a modeling choice motivated by the fact that the cost coefficient $c$ (the cost of the speculation phase) must be lowered by, for example, decreasing the latency of the drafter model. 

We express the size of the design space as $v \cdot N^{m}$, where $v$ is the number of design variants, $N$ the number of distinct \acp{PU} available for assignment, and $m$ is the number of subgraph partitions. In this formulation, a \textit{design variant} corresponds to a unique combination of cores, shaders, or PEs available across all \acp{PU}, which we count by placing its unordered configurations in bijective correspondence with the Cartesian product 
\begin{equation*}
v = \prod_{i} n_{i},
\end{equation*}

where $n_{i}$ denotes the number of available cores, shaders, or PEs in $\text{PU}_i$. For example, Fig.~\ref{fig:dse} (top) depicts a CPU with six physical cores and a GPU with a single shader that yields $v = 6 \times 1 = 6$ distinct hardware configurations for mapping subgraphs. In practice, only a subset of these resources may be accessible at runtime, for example two active CPU cores as shown in gray. Moreover, for the configuration illustrated in Fig.~\ref{fig:dse}, assuming two \acp{PU} available for mapping ($N = 2$) (top) and two graph partitions ($m = 2$, one for the drafter and one for the target, bottom), the size of the design space becomes $v = 6 \cdot 2^{2} = 24$ possible mappings.

% \vspace{0.3em}
% \begin{figure}[!t]
%     \centering
%     \includegraphics[width=0.9\linewidth]{figures/DSE.pdf}
%     \caption{Example of a design space with $v = 6$ design variants and two available \acp{PU} (left) and $m = 2$ subgraph partitions (right). The CPU provides six cores and the GPU a single shader, but in this example only two CPU cores (in grey) are assumed to be available for mapping.\jp{Give a label to the rect at the bottom. Memory? Graphical elements without meaning: 1) there is no need for a second line around the inner one, 2) the black rect on background has no meaning either and doesn't match the style adopted in other figs, 3) lines with shading also don't match the style of the paper. Wouldn't be better to merge the elements of this figure in Fig. 2? Fig. 3 seems pretty introductory as well and Fig. 2 has still some space. }}
%     \label{fig:dse}
% \end{figure}

This space grows quickly if any of its factors increase: finer-grained partitioning (larger $m$) increases the exponent, adding more types of \acp{PU} (larger $N$) raises the base, and incorporating more cores, shaders, and PEs (larger $v$) multiplies the size of the space. In practice, therefore, exhaustive exploration becomes infeasible on richer platforms; for this reason, and to keep the study focused, we deliberately keep $v$, $N$, and $m$ small and rely on the analytical cost model presented in \eqref{eq} to guide the search.

To evaluate each mapping, we find the $\gamma$ value that yields the highest acceleration $S$, for a given cost coefficient $c$, that encodes the speculation computational cost of a given mapping, and a given acceptance rate $\alpha$, that encodes the drafter model speculation quality. Both $\alpha$ and $c$ are measured empirically as described next.

\subsection{Measuring Acceptance Rates and Cost Coefficients}
\label{subsec:measurement}

We determine empirically the acceptance rate $\alpha$ using a 16-core server-grade CPU. Although $\alpha$ is a model-dependent value, reflecting how closely the drafter model approximates the larger target model, it remains hardware-independent\footnote{Slight deviations may occur when different devices handle rounding and precision differently, potentially leading to numerical discrepancies or precision divergence arising from quantization.}. Concretely, to estimate $\alpha$, we employed the Spec-Bench, a benchmark designed to evaluate \ac{SD} performance across multiple tasks~\cite{xia_unlocking_2024}. The dataset comprises 480 samples distributed among thirteen tasks.

In edge deployments, such task variability is lower because models are tailored to a single domain and typically benefit from knowledge distillation or task-specific fine-tuning. For this reason, we focus on the translation task. Additionally, due to the nature of translation, the length of the generated tokens tends to closely match the input sequence length, which is typically short (a few tens of tokens representing a sentence).

As quantization is a key enabler of ML systems on edge devices, we evaluated the acceptance rate $\alpha$ of multiple model configurations. These included:

\begin{itemize}
    \item fully quantized target-drafter pairs,
    \item semi-quantized combinations, and
    \item unquantized \texttt{FP16} counterparts as reference models.
\end{itemize}

All quantization schemes are static and implemented in \texttt{w8a8} precision using the Intel Neural Compressor~\cite{noauthor_intelneural-compressor_2025}.

% We do not consider a semiquantized configuration, where the target is quantized but the drafter remains unquantized, because the memory footprint during initialization (14.76 GB) would place a significant burden on system memory of the edge device. Conversely, the configuration (ii) has a with a total memory footprint of 5.66 GB for the model weights; therefore, the memory required during initialization ascends to 11.32 GB.  

We compute $c$ by compiling a single forward pass of both the drafter and target models using IREE~3.6.0, evaluating all combinations of the LLVM and SPIR-V backends independently, and benchmarking their respective runtimes. The ratio between these measured latencies directly yields $c$, thereby quantifying the relative computational cost of the speculation phase with respect to the target model for each homogeneous and heterogeneous configuration.

Given the measured acceptance rate $\alpha$ and $c$ values, the cost model given by \eqref{eq} is used to estimate the expected speedup and the optimal speculative length $\gamma$ for each device pairing (homogeneous and heterogeneous). We then calculate the best-performing mapping and then validate the predicted speedup experimentally on hardware, comparing the resulting acceleration against a CPU-only, non-speculative baseline.

\subsection{Matching Abstractions of Speculative Sampling and Heterogeneous Execution}
\label{subsec:abstractions}

Selecting the abstraction level for spatial partitioning onto heterogeneous \acp{PU} presents a tension: speculative sampling is most naturally expressed in high-level ML frameworks, whereas heterogeneous partitioning requires low-level control. We address this by rising hardware partitioning into the compiler frontend and integrating it directly within the speculative sampling pipeline. This unified, high-level formulation preserves the programmability of PyTorch-level code while enabling targeted acceleration (e.g., mapping the speculative phase onto an appropriate \ac{PU}).

% IREE, backed by \ac{MLIR}, enables mixing in the same \ac{IR} high-level dialects (e.g., Torch IR) with tensor semantics with lower-level dialects (e.g., IREE's Flow dialect), allowing to express spatial partitioning directly at the PyTorch level using custom operators that are later resolved to physical devices during progressive lowering. This unified abstraction supports injecting low-level data-flow operators into higher-level graphs and enables the entire speculative sampling pipeline, including both model arithmetic and procedural control flow, to be captured within a single optimized module, reducing glue code and supporting end-to-end optimization while preserving modularity through subgraph encapsulation; this approach is presented in Fig.~\ref{fig:monolithic_approach}. However, a fully monolithic \ac{AOT} compilation increases complexity, particularly since partial rewrites of the code are required to ensure compatibility with graph capture mechanisms such as tracing; this issue is particularly critical for procedural logic involving control flow depending on dynamic data. 

\begin{figure}[!h]
    \centering
    \includegraphics[width=0.95\linewidth]{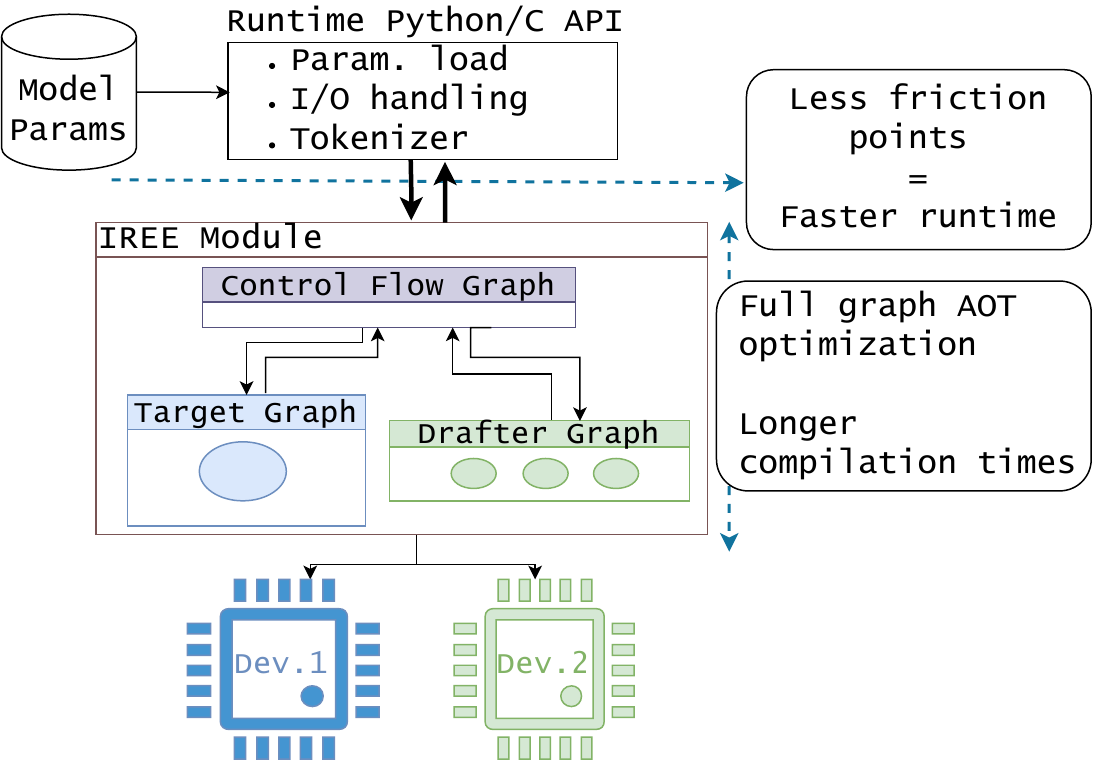}
    \caption{Monolithic approach: single IREE module with target, drafter, and control flow subgraphs with heterogeneous device affinities.
    % \gk{we need bigger fonts. At same time that is hard to read, we have a lot of empty space... }
    }
    \label{fig:monolithic_approach}
\end{figure}

IREE, backed by \ac{MLIR}, enables mixing high-level dialects (e.g., Torch IR) with lower-level dialects (e.g., Flow), allowing spatial partitioning to be expressed at the PyTorch level using custom operators resolved to physical devices during lowering. This enables capturing the entire pipeline within a single optimized module (Fig.~\ref{fig:monolithic_approach}), reducing glue code while supporting end-to-end optimization. However, monolithic \ac{AOT} compilation increases complexity, particularly for procedural logic with data-dependent control flow.

\begin{figure}[!h]
    \centering
    \includegraphics[width=0.95\linewidth]{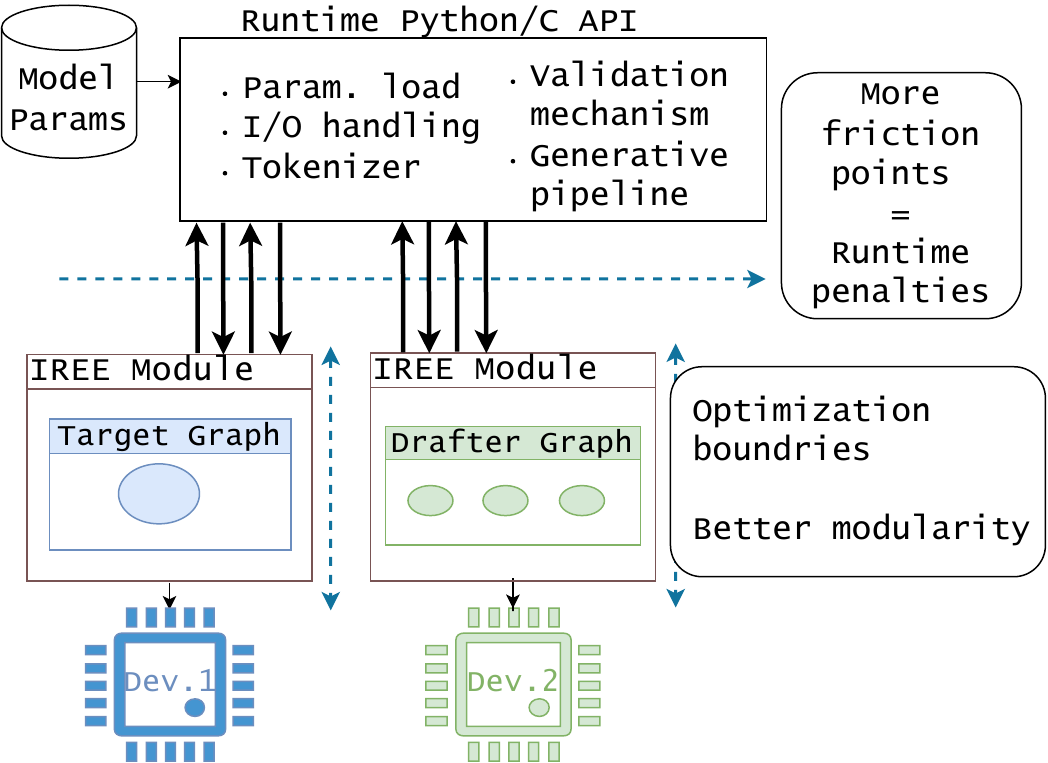}
    \caption{Modular approach: separate IREE modules for model arithmetic with control flow in the serving platform. 
    % \gk{same for this one. Bigger fonts }
    }
    \label{fig:separated_approach}
\end{figure}

% The approach illustrated in Fig.~\ref{fig:separated_approach} alleviates these challenges by adopting a modular design in which the target and drafter models are compiled independently, while the procedural logic is executed within the serving mechanism. This separation reduces recompilation latency and simplifies heterogeneous execution. However, it also introduces stricter boundaries between the model arithmetic modules and the control-flow logic. These boundaries are reflected by the larger number of thick black arrows in Fig.~\ref{fig:separated_approach}, each denoting an additional API call between the serving mechanism (top) and the underlying IREE modules (bottom). The resulting increase in API-call frequency introduces runtime overheads. We evaluate both design alternatives and report their performance characteristics in the following section.

Alternatively, Fig.~\ref{fig:separated_approach} shows a modular design where target and drafter are compiled independently while procedural logic executes in the serving mechanism. This simplifies heterogeneous execuiton but introduces stricter boundaries between modules, reflected by additional API calls (thick black arrows), which add runtime overhead. We evaluate both alternatives in the following section.

% \subsection{Execution Pipeline}
% The exploration pipeline thus consists of the following stages:  
% (i) compiling the drafter and target subgraphs with explicit device placement;  
% (ii) executing partitions on the chosen devices;  
% (iii) performing draft generation using the drafter on CPU or GPU;  
% (iv) validating drafted tokens on the target model; and  
% (v) advancing the decoding position based on acceptance outcomes.

% =====================================================================
\section{Experimental Setup and Evaluation}
\label{sec:evaluation}

Tab.~\ref{tab:exp_settings} summarizes the experimental configuration adopted in this work. It outlines the workload characteristics, task definition, optimization objective, \ac{SD} strategy, and the heterogeneous hardware resources considered during evaluation. In our setup, greedy sampling is used across all experiments and no KV cache is enabled. We employ the Llama~3.2 family, specifically Llama~3.2~1B as drafter and Llama~3.2~3B as target, following prior work demonstrating that alignment of the training data distributions between drafter and target is beneficial for draft acceptance~\cite{leviathan_fast_2023}. 
% JC: This paragraph+table is ill-placed here. Can't this be moved ahead of the evaluation as exp. setup? 

\begin{table}[!t]
\caption{Experimental settings}
\begin{center}
\scriptsize
\renewcommand{\arraystretch}{1.2}
\begin{tabular}{|c|c|}
\hline
\textbf{Setting} & \textbf{Description} \\
\hline
Workload & Short sequence lengths ($S_L \ll d$) \\
\hline
Task & Translation (from Spec-Bench benchmark~\cite{xia_unlocking_2024}) \\
\hline
Optimization objective & Minimize latency \\
\hline
\ac{SD} technique & Speculative sampling with target: Llama~3.2~3B \\ & and drafter: Llama~3.2~1B. \\
\hline
Heterogeneous edge hardware & Mali-G310 GPU and Hexacore Cortex-A55 \\ & CPU on NXP i.MX95 SoC \\
\hline
\end{tabular}
\label{tab:exp_settings}
\end{center}
\end{table}

\subsection{Empirical Estimation of Acceptance Rate}
\label{subsec:alpha_ev}

The acceptance rate $\alpha$ serves as a required input parameter for the analytical cost model given by \eqref{eq}. While improving $\alpha$ through better quantization schemes~\cite{zhang_speculative_2025} or training quantized drafters directly~\cite{li_eagle-3_2025} falls outside the scope of this work, we must characterize its behavior to validate our heterogeneous mapping methodology on real hardware. Evaluating these techniques on edge hardware represents a direction for future work; our contributions lies in how to exploit $\alpha$ effectively.

\begin{figure}[t]
    \centering
    \begin{subfigure}[b]{0.85\linewidth}
        \centering
        \includegraphics[width=\linewidth,trim={0 0.4cm 0 0},clip]{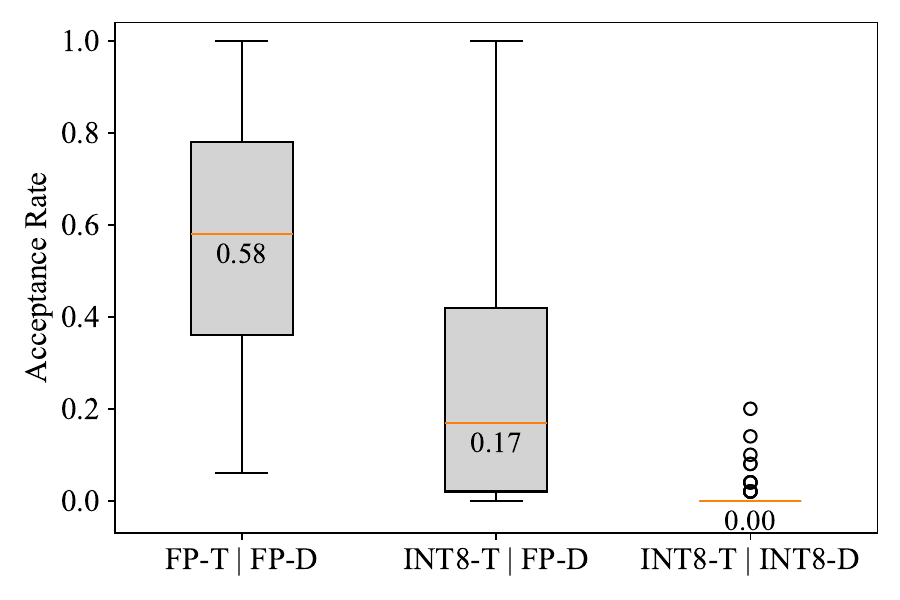}
        \caption{Translation acceptance rates comparison}
        \label{fig:acceptance_tranlation} 
    \end{subfigure}
    
    \begin{subfigure}[b]{0.85\linewidth}
        \centering
        \includegraphics[width=\linewidth,trim={0 0.4cm 0 0},clip]{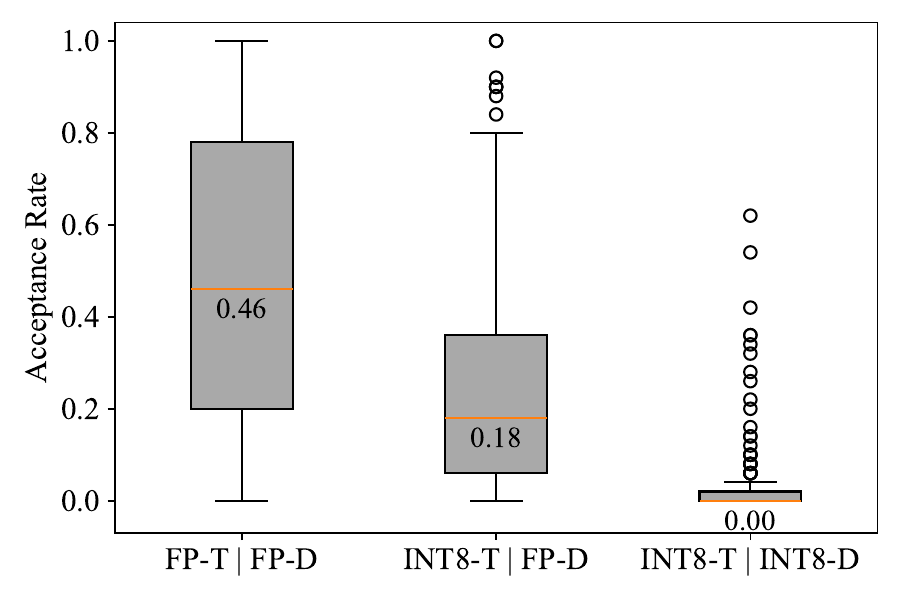}
        \caption{Entire dataset acceptance rates comparison}
        \label{fig:acceptance_all} 
    \end{subfigure}
    \caption{Acceptance rate $\alpha$ distribution for different quantization schemes: FP (FP32), T (target), D (drafter).
% \jp{x-axis labels are taking too much space. Either you add a linebreak or use a shorter label like FP-T/FP-D and INT-D/INT-T. In the caption, you say what T and D stand for. Export as PDF separetely.}
}
    \label{fig:accpentace_ratesy}
\end{figure}

Fig.~\ref{fig:accpentace_ratesy} shows the impact of quantization on $\alpha$ for the translation task (Fig.~\ref{fig:acceptance_tranlation}) and the full Spec-Bench dataset (Fig.~\ref{fig:acceptance_all}). The vertical axes represent the acceptance rate $\alpha$, while each box represents the distribution of $\alpha$ for three pairs of models: from left to right: an unquantized \texttt{FP16} drafter and target, a partially quantized configuration, where just the target model is quantized\footnote{Note that the fully unquantized configuration, as well as the partially quantized variant in which only the drafter is quantized, were excluded from the experiments due to memory limitations during initialization on our hardware-software setup.}, and a fully quantized \texttt{INT8} setup. As the level of quantization increases, the boxes shift downward, showing a consistent reduction in the median value for $\alpha$. In particular, the fully quantized pairing median collapses toward $\alpha \approx 0$, indicating that almost no drafted tokens are accepted. In contrast, the unquantized configuration achieves a median value of $\alpha = 0.58$. 

These results confirm prior work~\cite{zhang_speculative_2025}: as quantization increases, it introduces a distributional mismatch between drafter and target models, and the median $\alpha$ decreases substantially for speculative sampling on edge devices. This pattern is consistent across the entire Spec-Bench dataset, as shown in Fig.~\ref{fig:acceptance_all}, where the median falls along with increased quantization. The subsequent analysis concentrates on the translation task using the semiquantized configuration, for two reasons: (1) it provides a realistic edge deployment scenario balancing memory constraints with model accuracy, and (2) its broad distribution of acceptance rates, with $\alpha$ values spanning from $0\%$ to $100\%$, enables an exploration of the full range of speculative sampling behavior. 

\subsection{Computation of the Cost Coefficient}
\label{subsec:cost_coeff}
\begin{figure}[!h]
    \centering
    \begin{subfigure}[b]{0.85\linewidth}
        \centering
        \includegraphics[width=\linewidth,trim={0 0.6cm 0 0.5cm},clip]{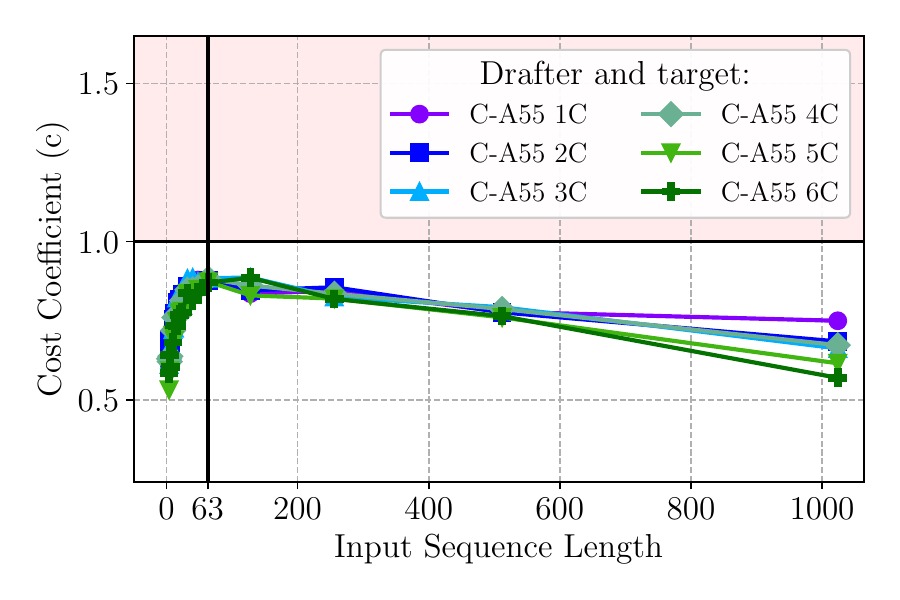}
        \caption{Measured cost coefficient (homogeneous execution)}
        \label{fig:cost_coefficient_homo} 
    \end{subfigure}

    \begin{subfigure}[b]{0.85\linewidth}
        \centering
        \includegraphics[width=\linewidth, trim={0 0.6cm 0 0},clip]{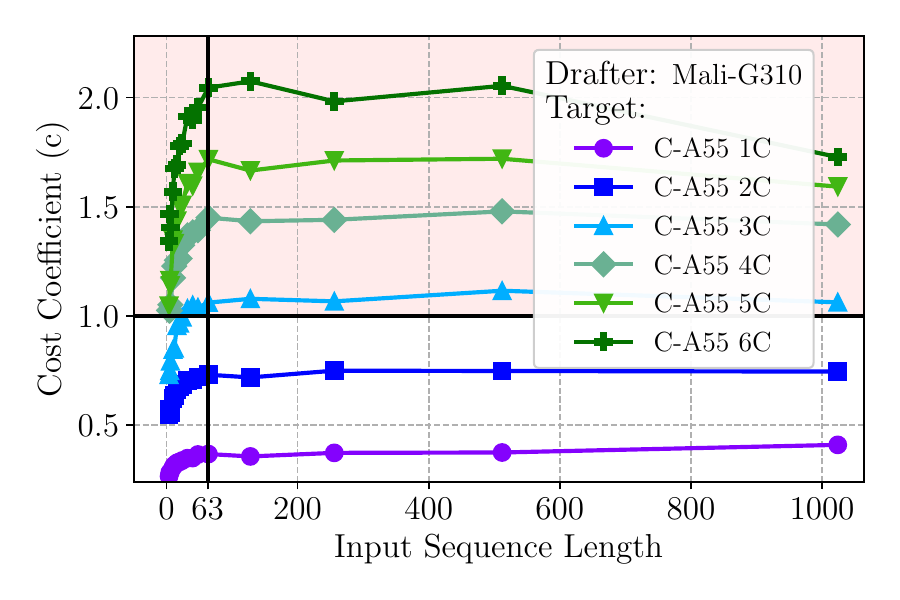}
        \caption{Measured cost coefficient (heterogeneous execution)
        % \jp{Data points are still almost impossible to see. Try to increase the opacity of the box or, prefebly, use symbolic legend (eg, C-A55-1C). The color of "4 cores" is very difficult to differentiate from the background, pls choose a color that has some contrast.}
        }
        \label{fig:cost_coefficient_hete} 
    \end{subfigure}
    \caption{Cost coefficients $c$ for homogeneous (\subref{fig:cost_coefficient_homo}) and heterogeneous (\subref{fig:cost_coefficient_hete}) mappings as a function of input sequence length. The black vertical line indicates $S_L=63$, the average input sequence length for the translation task in the Spec-Bench C‑A55 nC: Cortex‑A55 CPU, n cores. 
    % \jp{Use the same figure height vs. width proportion from Fig. 6. Fig.a) yaxis upper limit could be 1.5 and you could use 2 columns in the legend. Fig.b) Shrink text in legend. You can use "Target (Drafter: Mali-G310)" as the legend title and then reuse Cortex-A55 (X cores) legend as above. Most of the data points should be visible!}
    % \jp{Increase fontsize (axis ticks, title, labels). I'd rather have each plot exported as PDFs separately, and we can organize them as subfigures in the text.} 
    }
    \label{fig:cost_coefficient_comparison}
\end{figure}

We individually measured the latency of one forward pass for both the target and drafter models as a function of the input sequence length and then calculated the cost coefficient $c$, summarized in Fig.~\ref{fig:cost_coefficient_comparison}. The horizontal axes represent the input sequence length, and each curve depicts one design variant. Fig.~\ref{fig:cost_coefficient_homo} shows $c$ values for homogeneous mappings on CPU, whereas Fig.~\ref{fig:cost_coefficient_hete} shows heterogeneous mappings where the drafter is executed on the GPU\footnote{We do not map the quantized target model onto the GPU because it does not support \texttt{INT8} datatype; any \texttt{INT8} tensor is promoted to \texttt{FP32}, adding overhead and diminishing the benefits of quantization.}. The regions with $c > 1$ are shaded in red to indicate infeasible configurations in which one forward pass of the drafter is slower than the target model. These infeasible cases arise mainly when the system uses three to six CPU cores in the heterogeneous mapping (i.e., drafter model mapped on the GPU). 

A notable case appears in the design variant with only a single CPU core available (purple curves in both plots). When performing a heterogeneous mapping, the cost coefficient $c$ decreases substantially, from approximately $0.80$ (purple curve at the top for an input sequence length of $63$)\footnote{This input sequence length corresponds to the average length of the translation task in the Spec-Bench dataset.} to $0.41$ at the same sequence length, shown at the bottom. This improvement arises because the GPU executes the drafter model roughly three times faster than a single CPU core, significantly reducing the computational cost of the speculative phase. As a result, this configuration emerges as the most favorable candidate for speculative sampling on the evaluated SoC using the Spec-Bench dataset.

\subsection{Estimation of the Speedup with the Cost Model}
The expected speedup is estimated using Eq.~\eqref{eq} for two acceptance rates: a high value of $\alpha = 0.90$ corresponding to the 90th percentile and a low value of $\alpha = 0.17$ corresponding to the median. The results are reported in Tab.~\ref{tab:ttft_speedup_21_inputSeqLen} and Tab.~\ref{tab:ttft_no_speedup_21}, respectively. For $\alpha = 0.90$ (Tab.~\ref{tab:ttft_speedup_21_inputSeqLen}), speculative sampling combined with heterogeneous execution yields substantial improvements, particularly in design variant~1, which corresponds to a configuration with one CPU core available for mapping and the GPU (recall that a design variant represents a unique combination of cores, shaders, or PEs across all the \acp{PU}). This variant reaches a speedup of $1.68\times$ using $\gamma = 5$. The correct selection of the draft sequence length ($\gamma$) is crucial: each design variant achieves its best performance with a different $\gamma$ value, ranging from 0 (no speculative sampling) to 5. 

\begin{table}[!h]
\caption{Estimated speedup for $\alpha = 0.90$ and $S_L = 63$}
\begin{center}
\renewcommand{\arraystretch}{1.3}
\begin{tabular}{|c|c|c|c|}
\hline
\textbf{Design} & \textbf{Speculative} & \textbf{Heterogeneous} & \textbf{Speedup} \\
\textbf{Variant} & \textbf{Sampling} & \textbf{Execution} & \textbf{[$\times$]} \\
\hline
1 & \cellcolor{green!20}Yes ($\gamma = 5$) & \cellcolor{green!20}Yes & 1.68 \\
\hline
2 & \cellcolor{green!20}Yes ($\gamma = 2$) & \cellcolor{green!20}Yes & 1.10 \\
\hline
3 & \cellcolor{red!20}No & \cellcolor{gray!20}NA & 1 \\
\hline
4 & \cellcolor{red!20}No & \cellcolor{gray!20}NA & 1 \\
\hline
5 & \cellcolor{green!20}Yes ($\gamma = 1$) & \cellcolor{red!20}No & 1.02 \\
\hline
6 & \cellcolor{red!20}No & \cellcolor{gray!20}NA & 1 \\
\hline
\end{tabular}
\label{tab:ttft_speedup_21_inputSeqLen}
\end{center}
\end{table}

\begin{table}[!h]
\caption{Estimated speedup for $\alpha = 0.17$ and $S_L = 63$}
\begin{center}
\renewcommand{\arraystretch}{1.3}
\begin{tabular}{|c|c|c|c|}
\hline
\textbf{Design} & \textbf{Speculative} & \textbf{Heterogeneous} & \textbf{Speedup} \\
\textbf{Variant} & \textbf{Sampling} & \textbf{Execution} & \textbf{[$\times$]} \\
\hline
1--6 & \cellcolor{red!20}No & \cellcolor{gray!20}NA & 1 \\
\hline
\end{tabular}
\label{tab:ttft_no_speedup_21}
\end{center}
\end{table}

In contrast, design variants with a higher number of available cores (i.e., 3, 4, and 6 cores upwards) are expected to have a performance decline when both speculative sampling and heterogeneous execution are applied. Therefore, these methods should be avoided in such configurations. For the design variant holding five CPU cores, there is a modest speedup, but heterogeneous mapping should not be applied. If the performance gain is too small, there is a risk that, in a real system, the improvement becomes negligible, especially when accounting for deployment overheads. As a result, we discourage the use of heterogeneous mapping in this scenario. However, for $\alpha = 0.17$ (Tab.~\ref{tab:ttft_no_speedup_21}), even the shortest draft length $\gamma=1$ becomes detrimental and neither speculative sampling nor heterogeneous execution yields a speedup in any design variant.

\subsection{Validation and Discussion}
\label{sec:val_and_dis}

\begin{figure}[t!]
    \centering
    \begin{subfigure}[b]{0.85\linewidth}
        \centering
        \includegraphics[width=\linewidth,trim={0 0.6cm 0 1cm},clip]{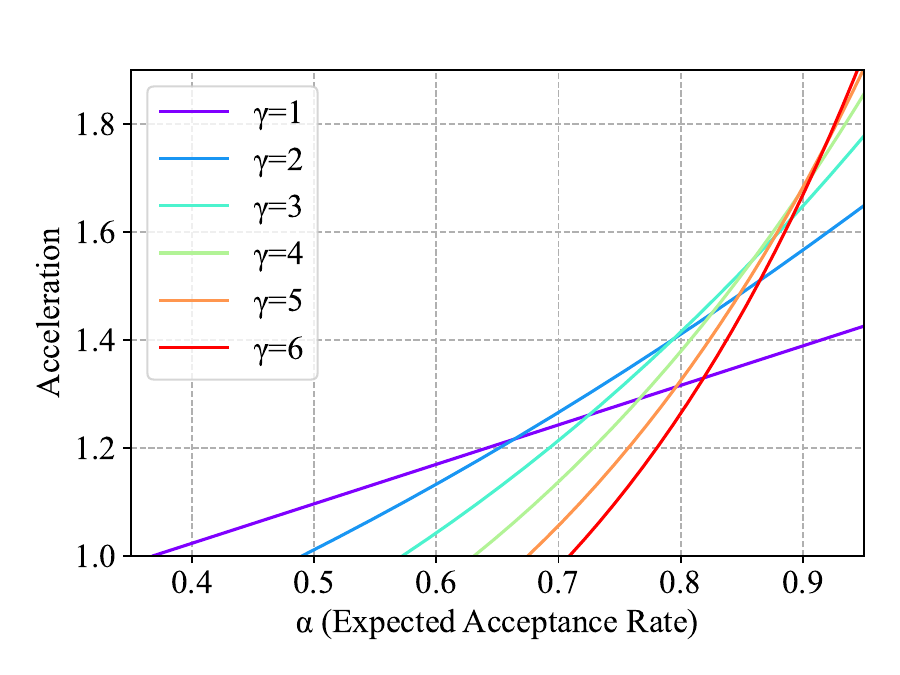}
        % \captionsetup{justification=centering, singlelinecheck=off}
        % \caption{Estimated acceleration for mapping:\\
        % Target model $\rightarrow$ Cortex-A55 1 core\\
        % Drafter model $\rightarrow$ Mali-G310\\
        % at input length $=63$} 
        \caption{Estimated acceleration}
        \label{fig:predicted_acceleration} 
    \end{subfigure}
    
    \begin{subfigure}[b]{0.85\linewidth}
        \centering
        \includegraphics[width=\linewidth,trim={0 0.6cm 0 0},clip]{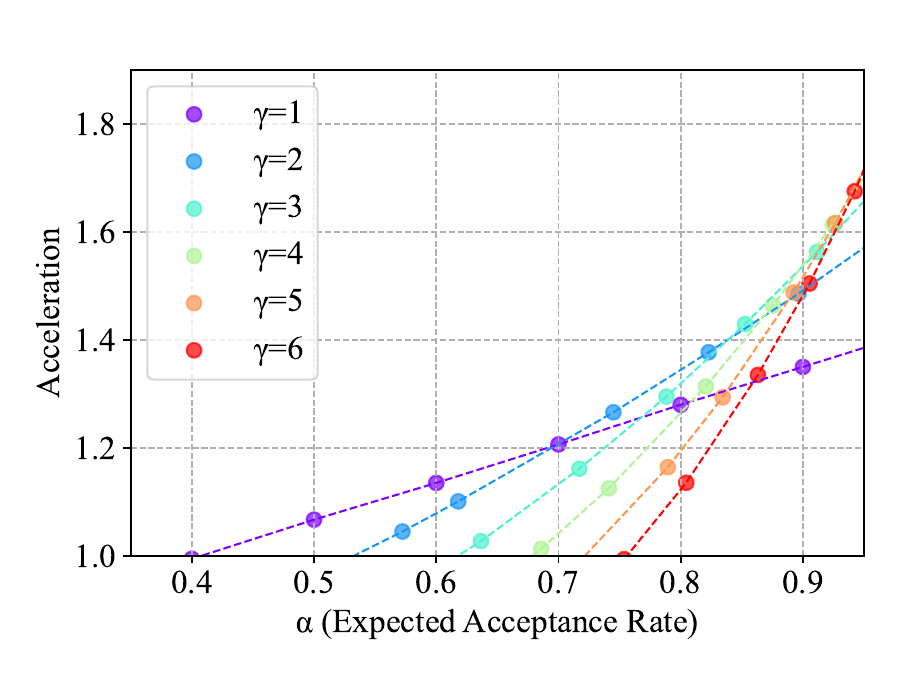}
        % \captionsetup{justification=centering, singlelinecheck=off}
        % \caption{Measured acceleration for mapping: \\Target model $\rightarrow$ Cortex-A55 1 core \\Drafter model $\rightarrow$ Mali-G310 \\at input length $=63$}
        \caption{Measured acceleration}
        \label{fig:measured_acceleration} 
    \end{subfigure}
	\caption[Measured acceleration using speculative sampling and heterogeneous execution]{Measured acceleration introduced by speculative sampling with different drafted token lengths $\gamma$ executed heterogeneously: Target model mapped on one CPU core; drafter model, on GPU. The input sequence length is set to 63 and we report the average after ten runs.
    % \jp{you can make these figures shorter.}
    }
	\label{fig:measured_and_predicted_acceleration}
\end{figure}

The underlying acceleration trend and its empirical validation are jointly illustrated in Fig.~\ref{fig:measured_and_predicted_acceleration}, where the horizontal axes represent the acceptance rate $\alpha$ and the vertical axes report the predicted or achieved acceleration $S$, in Figs.~\ref{fig:predicted_acceleration} and~\ref{fig:measured_acceleration}, respectively. In both cases, we present the heterogeneous mapping in which the quantized target model is executed on a single CPU core, while the unquantized drafter model is assigned to the GPU. The results for an input sequence length of 63 are shown. The different curves correspond to different draft lengths $\gamma$, with longer drafts (orange and red curves) offering greater acceleration at sufficiently high $\alpha$, while becoming detrimental when $\alpha$ falls. This explains why the semiquantized configuration with an acceptance rate of $\alpha = 0.17$ in Tab.~\ref{tab:ttft_no_speedup_21} yields no measurable speedup: the acceptance rate is simply too low for speculative sampling to be beneficial. Fig.~\ref{fig:measured_acceleration} presents the measured acceleration, showing that the expected acceleration (Fig.~\ref{fig:predicted_acceleration}) is achieved with an $\alpha$ approximately 4\% higher than estimated. Despite this shift, the empirical curve remains consistent with the analytic prediction.

% Importantly, the Python orchestration layer used to coordinate drafter and target execution does not introduce measurable overhead, likely because model computation dominates the overall runtime on the evaluated edge hardware.

The evaluated heterogeneous configuration (drafter on GPU, target on single CPU core) arises from hardware-software constraints (e.g., IREE lacking \texttt{INT8} support for Mali backends, quantization effects on $\alpha$ leading to semiquantized setups). Alternative strategies (full-GPU execution, dynamic scheduling) either exceed the memory budget of the platform, damage drafter accuracy, or require unavailable runtime capabilities. The analytical cost model (Eq.~\ref{eq}) abstracts hardware characteristics via $c$ to determine when heterogeneous execution is beneficial. The methodology generalizes to other platforms through profiling $c$ and measuring task-specific $\alpha$.

Runtime constraints in IREE 3.6.0 prevented deploying a monolithic graph with heterogeneous device affinities. Consequently, we compiled separate graphs (Fig.~\ref{fig:separated_approach}) orchestrated via IREE's runtime API. This introduces interface overhead that may explain the measured $4\%$ deviation, but provided necessary flexibility for heterogeneous execution.

% Nonetheless, the drafter phase remained monolithic, with separate compiled versions for each $\gamma$ value. Thus, in the speculative sampling pipeline, each generation cycle involved only two API calls: one to the appropriate drafter model and one to the target model. In the baseline case, only a single call to the target model was required per generation cycle. Despite the overheads, these results provide useful guidance for future heterogeneous deployments and point to the need for runtime systems that better support coarse-grained spatial partitioning.

% =====================================================================
\section{Conclusion and Outlook}
% =====================================================================
This work showed how compiler-assisted heterogeneous mapping can accelerate LLM inference with speculative sampling on resource-constrained edge devices. An analytical cost model provided a basis for deciding when speculative sampling and heterogeneous execution are worthwhile under fixed hardware budgets, and model validation showed a 4\% deviation from predictions. Exposing low-level abstractions into the compilation flow facilitated combining speculative sampling with heterogeneous execution, improving productivity with minimal performance penalties. Under favorable conditions (with a predicted $\alpha = 0.90$ and measured $\alpha=0.94$), Llama~3.2~3B achieved up to a 1.68$\times$ speedup. 

Future work should: (1) integrate improved quantization techniques to achieve higher $\alpha$ values; (2) validate the cost model with additional edge SoCs and other \ac{SD} techniques, as discussed in Sec.~\ref{subsec:workflow_overview}; (3) extend the model to incorporate finer-grained partitioning would allow mapping computation onto NPUs and other accelerators with restricted operation support; and (4) improve the runtime memory allocator for heterogeneous execution of monolithic graphs with mixed device affinities to allow shared GPU-CPU scheduling.

\section*{Acknowledgment}

% The preferred spelling of the word ``acknowledgment'' in America is without 
% an ``e'' after the ``g''. Avoid the stilted expression ``one of us (R. B. 
% G.) thanks $\ldots$''. Instead, try ``R. B. G. thanks$\ldots$''. Put sponsor 
% acknowledgments in the unnumbered footnote on the first page.
This result is part of the IPCEI ME/CT and is funded by the European Union Next Generation EU, the German Federal Ministry for Economic Affairs and Energy, the Bavarian Ministry of Economic Affairs, Regional Development and Energy, the Free State of Saxony with the help of tax revenue based on the budget approved by the Saxon State parliament and the Free and Hanseatic City of Hamburg. The authors acknowledge the financial support by the Federal Ministry of Research, Technology and Space of Germany and by Sächsische Staatsministerium für Wissenschaft, Kultur und Tourismus in the program Center of Excellence for AI-research ``Center for Scalable Data Analytics and Artificial Intelligence Dresden/Leipzig'', project identification number: ScaDS.AI.

\end{document}